%% file: paper.tex
\documentclass[]{TEAI}
\usepackage{helvet}
\usepackage{amsmath}
\usepackage{amssymb}
\usepackage{amsfonts}
\usepackage{graphicx}
\usepackage{subcaption}
\usepackage[toc,page,header]{appendix}
\usepackage[utf8]{inputenc}
\usepackage[T1]{fontenc}
\usepackage{hyperref}
\usepackage{url}
\usepackage{booktabs}
\usepackage{nicefrac}
\usepackage{microtype}
\usepackage[dvipsnames]{xcolor}
\usepackage{wrapfig}
\usepackage{float}
\usepackage{multirow}
\usepackage{makecell}
\usepackage{siunitx}
\usepackage{tikz}
\usepackage{pgfplots}
\usepackage{pgf-pie}
\usepackage[export]{adjustbox}
\usepackage{ragged2e}
\usepackage{tabularx}
\usepackage{array}
\usepackage{caption}
\usepackage{enumitem}
\usepackage{pifont}
\usepackage[hang,flushmargin]{footmisc}
\usepackage{tcolorbox}
\tcbuselibrary{breakable}
\tcbuselibrary{skins}
\usepackage{listings}
\usepackage{algorithm}
\usepackage{algpseudocode}
\usepackage{bbm}

\definecolor{lightblue}{RGB}{200, 230, 255}
\definecolor{headerblue}{RGB}{150, 200, 255}

\title{What Do LLM Agents Know About Their World? Task2Quiz: A Paradigm for Studying Environment Understanding}

\author{
  Siyuan Liu\textsuperscript{1},
  Hongbang Yuan\textsuperscript{1},
  Xinze Li\textsuperscript{1},
  Ziyue Zhu\textsuperscript{1},
  Yixin Cao\textsuperscript{1},
  Yu-Gang Jiang\textsuperscript{1}
}

\affiliation[1]{School of Computer Science, Fudan University}
\vspace{-20pt}

\abstract{
Large language model (LLM) agents have demonstrated remarkable capabilities in complex decision-making and tool-use tasks, yet their ability to generalize across varying environments remains a under-examined concern. Current evaluation paradigms predominantly rely on trajectory-based metrics that measure task success, while failing to assess whether agents possess a grounded, transferable model of the environment. To address this gap, we propose Task-to-Quiz (T2Q), a deterministic and automated evaluation paradigm designed to decouple task execution from world-state understanding. We instantiate this paradigm in T2QBench, a suite comprising 30 environments and 1,967 grounded QA pairs across multiple difficulty levels. Our extensive experiments reveal that task success is often a poor proxy for environment understanding, and that current memory machanism can not effectively help agents acquire a grounded model of the environment. These findings identify proactive exploration and fine-grained state representation as primary bottlenecks, offering a robust foundation for developing more generalizable autonomous agents.
}
\correspondence{\email{email@domain}}
\vspace{-10pt}
\begin{document}
\maketitle

\input{section/1_introduction}
\input{section/2_data_pipline}
\input{section/3_evaluation}
\input{section/4_experiment}

\input{section/5_relatedwork}
\input{section/6_conclusion}
\input{section/7_limitations}

\clearpage
\bibliographystyle{plainnat}
\bibliography{main}

\clearpage
\appendix
\input{section/8_appendix}

\end{document}

%% file: section/1_introduction.tex
\vspace{-20pt}

\section{Introduction}
\label{sec:intro}

Large language model (LLM) agents have recently shown impressive competence in realistic, multi-step decision making and tool use~\cite{yehudai2025surveyevaluationllmbasedagents,luo2025largelanguagemodelagent,matarazzo2025surveylargelanguagemodels,minaee2025largelanguagemodelssurvey}, enabling progress in web navigation~\cite{wei2025webagentr1trainingwebagents}, software engineering~\cite{zhang2024codeagentenhancingcodegeneration,dong2025surveycodegenerationllmbased}, and operating-system interaction~\cite{hu2025osagentssurveymllmbased}. While, the issue of generalization is concerned by the community. Agent capabilities that look strong on particular scenarios often fail to transfer reliably when the environment changes or constraints are slightly modified\cite{liu2025realbarrierllmagent}. This gap raises a basic but under-examined question: do LLM agents actually acquire a grounded model of the environment, or do they primarily learn specific heuristics that optimize task metrics?

\begin{wrapfigure}{r}{0.45\textwidth}
\centering
\includegraphics[trim=0cm 4cm 1.2cm 0cm, clip, width=\linewidth, page=2]{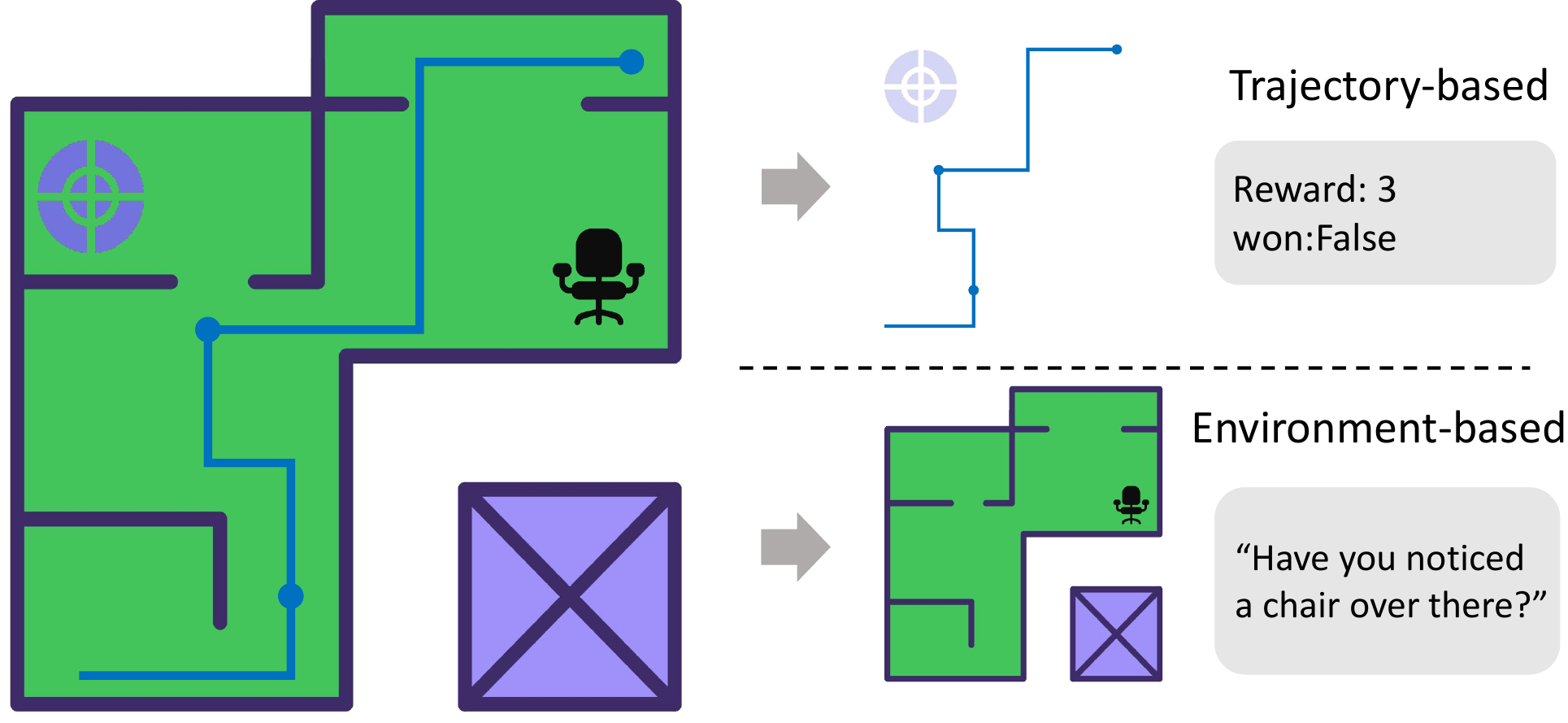}
\caption{Comparison of trajectory-based and environment-based Evaluation, the trajectory-based evaluation cares about whether and how the final goal is reached, while the environment-based evaluation cares about the entire understanding of the environment.}
\label{fig:difference}
\end{wrapfigure}

In this paper, we aim at investigating agent's environment understanding beyond current task success evaluation. As shown in Figure~\ref{fig:difference}, most existing agent benchmarks are trajectory-based. They ask whether an agent reaches the goal, such as success rate or the quality of intermediate actions, measuring ``doing''.
However, generalization or transferability requires ``knowing''.
An agent may complete a task while still lacking robust world-state knowledge, such as where objects are, how rooms connect, which direction relationships hold, or what latent object states (e.g., locked/closed) are. Conversely, an agent may acquire correct environment facts yet fail due to planning errors or long-horizon constraints. 
Therefore, we highlight environment-based evaluation to provide diagnostic ``knowing'' signals, support controlled interventions (e.g., varying layout complexity or reachability constraints), and enables reproducible grading grounded in known world metadata rather than human annotation or stochastic LLM-as-judge scoring. However, there are three challenges: (i) how to encourage agents to explore beyond single-goal exploitation; (ii) how to quantify environment understanding in a fine-grained, comparable way; and (iii) how to build a fully automated and reproducible pipeline without manual labeling or judge hallucinations~\cite{kalai2025languagemodelshallucinate}.

To address the issues, we propose \textbf{Task-to-Quiz (T2Q)}, an automated environment-based evaluation paradigm. We instantiate T2Q on TextWorld-style text games, where the full environment metadata (topology, entity placements, and symbolic state relations) is available by construction, making verification precise and convenient. There are two stages. Stage 1 synthesizes coverage-oriented task sets that drive agents to traverse rooms, interact with objects, and reveal stateful constraints; task generation is designed to increase exposure to diverse environment elements rather than only optimizing a single end goal. Stage 2 converts the environment and the executed trajectories into a multi-dimensional quiz that probes complementary aspects of world-state knowledge, including localization, connectivity, direction, key--lock matching, and latent properties. Importantly, each question is paired with trajectory-based prerequisites: if an agent`s interactions could not have revealed the necessary evidence (e.g., it never visited a room or never opened a container), the question is marked non-answerable to avoid penalizing agents for information they had no opportunity to observe. Because answers are computed by a verifier grounded in environment metadata and interaction logs, the entire evaluation is deterministic and reproducible. This design separates failures of exploration/exposure from failures of state tracking and reasoning, enabling a fair and diagnostic measure of world-state knowledge.

Using this paradigm, we build T2QBench, a controlled evaluation suite spanning three difficulty levels with 30 environments, 224 coverage-oriented tasks, and 1,967 environment-grounded QA pairs. We evaluate diverse proprietary and open-source models under a unified protocol that jointly measures two complementary dimensions: Task Success Rate (TSR) and Environment Understanding Score (EUS), capturing post-interaction world-state knowledge. Across extensive experiments with different memory mechanisms, we obtain three consistent insights.
First, task success is indeed not a reliable proxy for environment understanding. TSR and EUS can diverge as difficulty increases.
Second, existing memory systems do not consistently improve environment understanding; in many cases, a naive in-context baseline matches or outperforms memory-augmented agents, suggesting that current memory pipelines may lose fine-grained evidence.
Third, a low propensity for proactive exploration is a dominant bottleneck. Agents perform relatively better on questions answerable via short-term recall, but struggle on aspects that require actively uncovering latent properties or relations.
Together, these results suggest that improving generalization may require not only better planning and retrieval, but also mechanisms that explicitly support world-state acquisition and representation.

\begin{figure*}
\centering
\includegraphics[trim=0cm 5cm 0cm 0cm, clip, width=\linewidth, page=13]{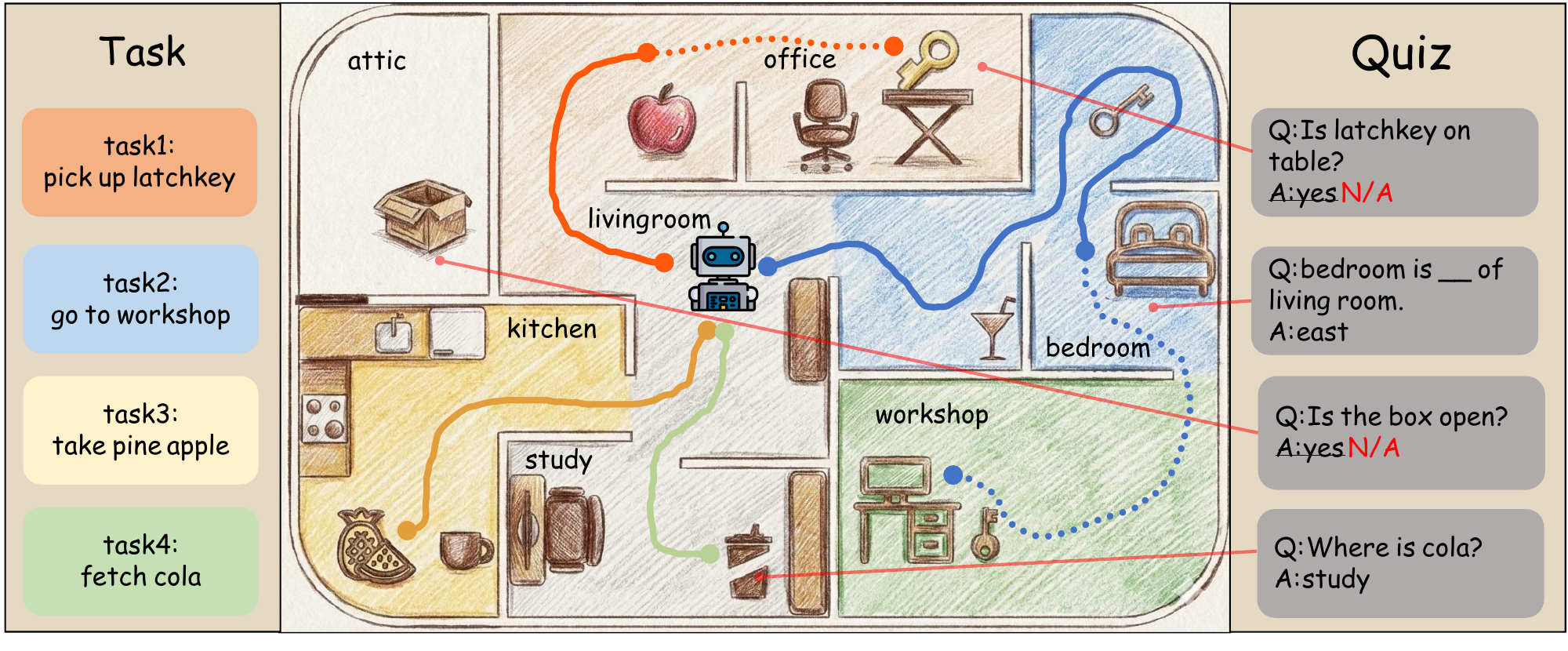}
\caption{Overview of our T2Q paradigm. It contains coverage-oriented tasks, and environment-based quizzes that are used to evaluate the agent's environment understanding. The answer of each question is dynamically generated based on the agent's interaction history and the environment metadata.}
\label{fig:T2QBench}
\end{figure*}

Our contributes can be summarized as follows:
\begin{itemize}
\item We propose a deterministic environment-based evaluation paradigm \textbf{Task-to-Quiz (T2Q)}, that separates ``doing'' from ``knowing''. T2Q contains two stages: 1) coverage-oriented task generation and 2) quiz evaluation, whose answer is dynamically generated based on the agent's interaction history and the environment metadata.
\item We construct a fully automated pipeline and build \textbf{T2QBench}, enabling reproducible measurement on different environments, tasks, and difficulty levels.
\item We conduct an empirical investigation across models and memory systems that reveals key bottlenecks in agents' environment understanding ability.
\end{itemize}

%% file: section/2_data_pipline.tex
\vspace{-15pt}
\section{Controllable Environment Construction}
In this section, we describe our controllable environment construction pipeline. We first present the base environment generation procedure, and then introduce the task coverage planning pipeline.
\subsection{Environment Generation}
\label{sec:environment-generation}
\paragraph{Layout and Connectivity Generation}We build the base game environments on top of TextWorld Framework~\cite{cote2019textworld} with a controllable configuration space. Specifically, we procedurally generate room layouts and connectivity and instantiate an explicit number of interactive objects. As a result, the full environment metadata is known by construction (e.g., connectivity, locations, and object attributes) for future verification and question generation.
\paragraph{Reachability Design}
The default TextWorld generation pipeline provides limited control over reachability because locks and keys are often sampled randomly, which may lead to weak coupling between access constraints and traversal structure. To make reachability controllable and systematically challenging, we add a post-generation refinement step that injects lock--key dependencies while preserving solvability: we first compute the reachable region from the start, then lock a fixed fraction of doors or containers (0.4 at our default setting) to enforce multi-step, long-horizon planning.
\paragraph{Distractor Placement}
To further increase reasoning difficulty without altering overall solvability, distractors are added during world construction. These include irrelevant objects and keys that are interactable but useless for task completion.

\subsection{Task Coverage Planning}

\paragraph{Modeling Task Generation as Weighted Set Cover Problem}
To comprehensively evaluate an agent's environment understanding capability, we aim to maximize its opportunities to interact with the environment. Since the exploration stage is task-driven, we design a task set that covers as much of the reachable area and interactive entities as possible while minimizing redundancy among tasks. Therefore, task generation is formulated as a \textbf{weighted set cover problem} (Algorithm~\ref{alg:Task_generation}). First, all reachable rooms and interactive entities are extracted from the environment instance $\mathcal{G}$ to form the target universe $\mathcal{R}$. They are considered as the targets that final task set should cover.
For each target in $\mathcal{R}$, a corresponding target-specific candidate task is instantiated.
To accomplish each task goal, the agent must execute a sequence of actions; we refer to the corresponding reference solution trajectory as a \textbf{walkthrough}.
Accordingly, a winning walkthrough is derived via TextWorld task analysis function using available metadata $\mathcal{K}$, producing a candidate task set $\mathcal{S}$.
These generated tasks serve as the candidate pool for the subsequent coverage selection step.

\begin{wrapfigure}{l}{0.55\textwidth}
    \vspace{-5pt}
    \small
    \captionsetup{type=algorithm,justification=raggedright,singlelinecheck=false}
    \begin{tcolorbox}[
      colback=black!5,
      colframe=black!70,
      boxrule=0.6pt,
      arc=1mm,
      left=2mm,right=2mm,top=1.5mm,bottom=1.5mm,
      before skip=0pt,
      after skip=0pt
    ]
      \captionof{algorithm}{Coverage Task Generation}
      \label{alg:Task_generation}
      \begin{algorithmic}[1]
        \Require Environment instance $\mathcal{G}$, metadata or knowledge $\mathcal{K}$
        \Ensure Greedy task set $\mathcal{Q}$
        \State $\mathcal{R} \leftarrow \textsc{ExtractTargets}(\mathcal{G})$
        \State $\mathcal{S} \leftarrow \textsc{PlanPaths}(\mathcal{G}, \mathcal{K})$
        \State $\mathcal{Q} \leftarrow \emptyset$
        \State $\mathcal{U} \leftarrow \mathcal{R}$
        \While{$\mathcal{U} \neq \emptyset$}
          \State $s^* \leftarrow \text{null}$, $g_{max} \leftarrow 0$
          \For{$s \in \mathcal{S} \setminus \mathcal{Q}$}
            \State $\Delta \leftarrow \textsc{Cover}(s) \cap \mathcal{U}$
            \State $g \leftarrow \textsc{Gain}(\Delta)$
            \If{$g > g_{max}$}
              \State $g_{max} \leftarrow g$
              \State $s^* \leftarrow s$
            \EndIf
          \EndFor
          \If{$s^*$ is null} \State \textbf{break} \EndIf
          \State $\mathcal{Q} \leftarrow \mathcal{Q} \cup \{s^*\}$
          \State $\mathcal{U} \leftarrow \mathcal{U} \setminus \textsc{Cover}(s^*)$
        \EndWhile
        \State \Return $\textsc{ToTasks}(\mathcal{Q})$
      \end{algorithmic}
    \end{tcolorbox}
    \vspace{-10pt}
\end{wrapfigure}

\paragraph{Coverage Signature}
When executing a task, an agent necessarily visits a subset of rooms and interacts with a subset of entities; we treat these visited and exercised elements as the task's \textbf{coverage signature}. For example, the task ``fetch an apple from the kitchen'' may require traversing the living room and bedroom and opening a chest, yielding a signature that covers \{living room, bedroom, chest\}. Formally, each candidate task $s \in \mathcal{S}$ is associated with a coverage signature $\textsc{Cover}(s) \subseteq \mathcal{R}$, which includes both the rooms traversed along the path and the interaction-relevant entities exercised by the task. Therefore, selecting a compact subset of tasks whose union covers $\mathcal{R}$ naturally induces a weighted set cover formulation.

\paragraph{Greedy Selection}

To improve data efficiency, we aim to minimize redundancy across tasks and keep the selected task set as compact as possible. We therefore adopt a greedy selection strategy that, at each step, chooses the task that contributes the largest marginal coverage over the yet-uncovered targets. A greedy weighted selection procedure is adopted in practice. At each iteration, the remaining uncovered set $\mathcal{U}$ is maintained, the marginal coverage is computed as $\Delta = \textsc{Cover}(s)\cap \mathcal{U}$. Then, the task maximizing a type-weighted gain $\textsc{Gain}(\Delta)$ is selected. The gain function assigns larger weights to \textbf{interactions} ($w_i$) and \textbf{objects} ($w_o$) than \textbf{rooms} ($w_r$) in a type-wise manner, prioritizing tasks that exercise complex behaviors over pure navigation. The procedure iterates until $\mathcal{U}=\emptyset$, yielding a minimal approximate task set $Q_{\min}$ with full target coverage and substantially reduced redundancy.

\subsection{Quiz Generation}
Beyond coverage-oriented tasks, a complementary quiz set is constructed to probe fine-grained grounded understanding of the environment. Question--answer pairs are generated from multiple perspectives.

\paragraph{Convert Environment Facts to Quiz Questions}
To automatically construct a comprehensive quiz suite that reflects an agent's understanding of and responsiveness to the environment, we generate quiz questions using deterministic rules grounded in the environment specification. Concretely, each environment instance is defined by a set of symbolic \textbf{facts} that encode fine-grained details such as entity locations, object attributes, room orientations, and connectivity (see Section~\ref{sec:detail-of-constructing-an-environment}). We extract a fact list that covers all such environment details and convert each fact into one or more question--answer pairs. For example, the fact $\texttt{in(apple, kitchen)}$ can be transformed into questions like ``Where is the apple?'' and ``Is the apple in the kitchen?''. This process yields a diverse quiz set that covers environment details across multiple question types.

\paragraph{Prerequisite Checkpoints}\label{sec:prerequisite-checkpoints}
In order to construct trajectory-conditioned, dynamic answers, prerequisite checkpoints are generated together with the questions. These checkpoints specify minimal interaction evidence required for an agent to be expected to know the answer. For example, a location question such as ``the apple is in the kitchen'' is only marked answerable if the agent has reached the kitchen.

\paragraph{Dynamic Ground Truth Generation}
Answers are assigned by combining static environment facts with the agent's interaction history. As Figure~\ref{fig:pipeline} shows, if all prerequisite checkpoints are satisfied, the agent is considered to have had sufficient opportunity to acquire the relevant knowledge, and the reference answer is used as the ground truth, which is regarded as ``\textbf{should have known}'' by the agent. Otherwise, the question is labeled as \textbf{non-answerable} for that trajectory to avoid penalizing agents for information they could not have observed. This hybrid of static answers and trajectory-conditioned answerability reached a balance between fairness and dynamic, behavior-aware evaluation.

%% file: section/3_evaluation.tex
\section{Evaluation}
\subsection{Dataset}

\begin{figure*}
\centering
\includegraphics[trim=0cm 4cm 4.5cm 0cm, clip, width=\linewidth, page=15]{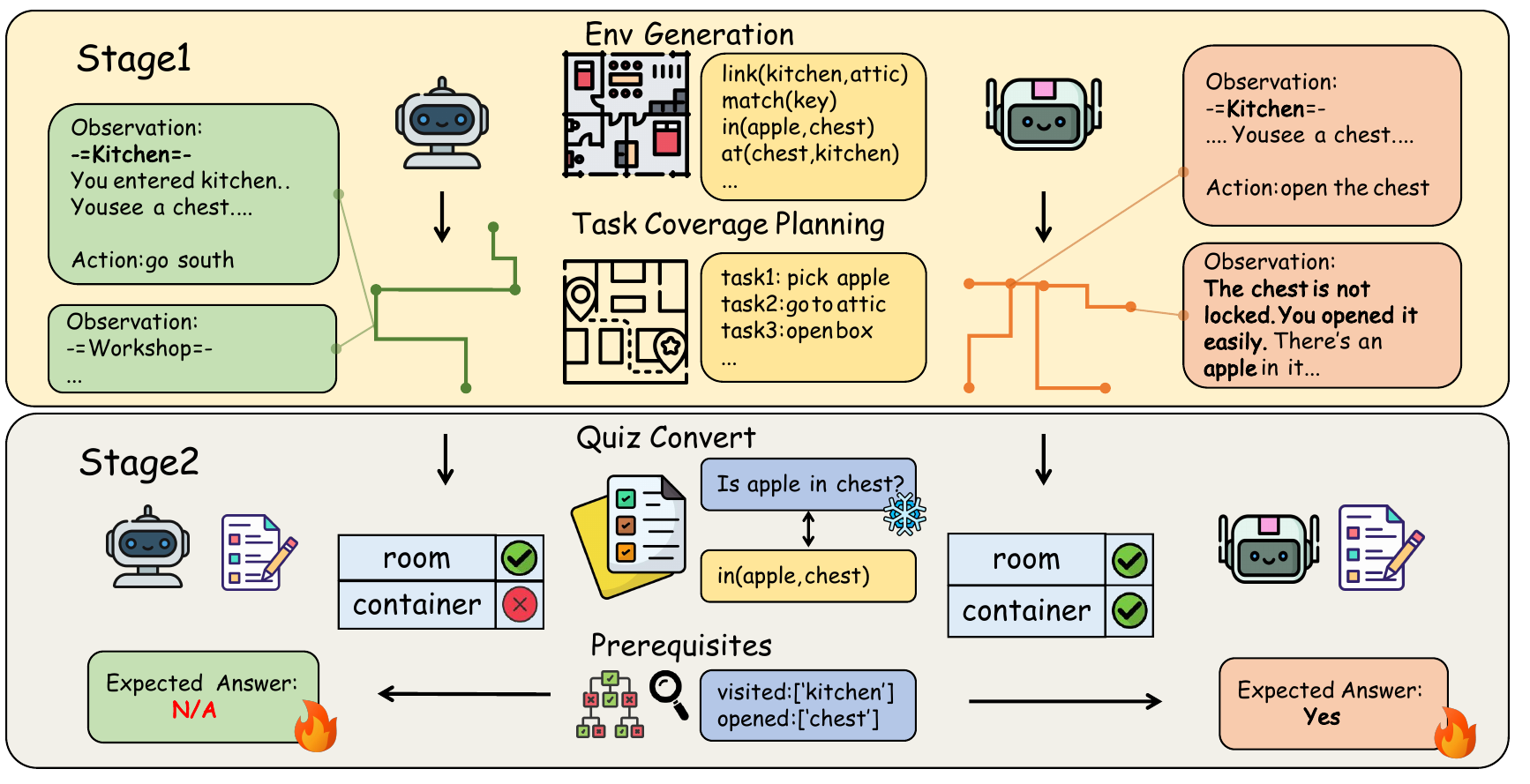}
\caption{Pipeline of data construction and 2-stage evaluation. Data is generated in deterministic rules during multiple steps, such as room layout, object placement, task coverage planning, and quiz generation. Agent needs to complete a set of tasks first. Then, the agent's interaction history and the environment metadata are used to generate the quiz set. Finally, the agent is asked to answer the quiz set based on the their memory on stage 1.}
\label{fig:pipeline}
\end{figure*}

\begin{wraptable}[9]{r}{0.40\textwidth}
    \vspace{-15pt}
    \centering
    \small
    \setlength{\tabcolsep}{4pt}
    \caption{Dataset Statistics for T2QBench. For each difficulty level, we generate 10 environments with varying numbers of rooms and objects.}
    \label{tab:dataset-statistics}
    \begin{tabular}{l c c c}
    \toprule
    \textbf{Setting} & \textbf{Easy} & \textbf{Medium} & \textbf{Hard} \\
    \midrule
    Rooms & 3--5 & 6--10 & 16--20 \\
    Objects & 6--10 & 14--18 & 28--32 \\
    Avg. Tasks & 2.4 & 5.7 & 12.3 \\
    Avg. Walkthrough & 4.17 & 5.25 & 5.78 \\
    Avg. Questions & 34.6 & 65.6 & 94.7 \\
    \bottomrule
    \end{tabular}
    \vspace{-10pt}
\end{wraptable}

\paragraph{T2QBench}We construct \textbf{T2QBench} via our pipeline and stratify environments into three difficulty levels based on the number of rooms and interactive objects. For each difficulty, we generate 10 distinct environments. Each environment is paired with a coverage-oriented task set and a quiz set. In total, the benchmark comprises 30 environments, 224 tasks, and 1,967 questions.

\paragraph{Quiz Taxonomy}
We categorize quiz questions into five types to probe complementary facets of grounded understanding:
\begin{itemize}
\item \textbf{Location (Loc.)}: asks where an entity is situated (e.g., the room or container of an object), which is often answerable given direct observation evidence.
\item \textbf{Connectivity (Conn.)}: queries the existence of traversable links between locations under the environment topology.
\item \textbf{Direction (Dir.)}: tests spatial orientation between locations typically demanding cross-turn aggregation and consistent coordinate reasoning.
\item \textbf{Match (Match.)}: tests the agent's ability to determine which key is the correct one for the lock.
\item \textbf{Property (Prop.)}: concerns latent object states, such as locked or unlocked, open or closed. Agent needs to interact with the object to reveal the property. For example, agent can only reveal a box is locked after trying to open it.
\end{itemize}
This taxonomy aligns generation with evaluation: each question is paired with trajectory-dependent prerequisites as described in Section~\ref{sec:prerequisite-checkpoints} so that type-wise accuracy reflects reasoning or memory given sufficient exposure rather than missing exploration.

\subsection{Formulation of Metrics}
\paragraph{Stage 1: Task Success Rate}
\noindent
We evaluate an agent's performance on the generated task set, reporting TSR.
Given an environment $\mathcal{E}$, we instantiate a task set
$\mathcal{T}=\{\tau_1,\tau_2,\dots,\tau_{|\mathcal{T}|}\}$ by the coverage-oriented task generation logic as Algorithm~\ref{alg:Task_generation}.
We then evaluate an agent (policy) $\pi$ by executing each task $\tau_i$ in $\mathcal{E}$.
The resulting interaction trace is a trajectory
$\zeta_i=(s_0,a_0,s_1,a_1,\dots,s_{T_i})$, and we denote the set of trajectories as
$\mathcal{Z}=\{\zeta_1,\zeta_2,\dots,\zeta_{|\mathcal{T}|}\}$.
We define a task success indicator as $\mathbbm{I}_{\mathrm{win}}(\zeta_i,\tau_i)\in\{0,1\}$, and report the task success rate:
\begin{equation}
\vspace{-5pt}
\mathrm{TSR}
:=\frac{1}{|\mathcal{T}|}\sum_{i=1}^{|\mathcal{T}|}\mathbbm{I}_{\mathrm{win}}(\zeta_i,\tau_i).
\end{equation}
where $|\zeta_i|$ is the number of actions in $\zeta_i$, and $\ell^{*}(\tau_i)$ is the oracle shortest action length to complete $\tau_i$ in $\mathcal{E}$.
For memory-augmented agents, Stage~1 interactions also update an internal memory state; we denote the resulting memory after executing the task set (associated with $\mathcal{Z}$) as $M_{\mathcal{Z}}$, which is carried into Stage~2.

\vspace{-5pt}
\paragraph{Stage 2: Environment Understanding Score}
\noindent
In Stage~2, we assess grounded understanding with quizzes spanning orientation, connectivity, object properties, and entity localization; importantly, a subset of questions depends on the agent's realized trajectory and thus requires interaction-conditioned reference answers.
We further instantiate a question set
$\mathcal{Q}=\{q_1,q_2,\dots,q_{|\mathcal{Q}|}\}$ to probe the agent's grounded understanding of $\mathcal{E}$.
Crucially, many questions are \textbf{trajectory-dependent}; therefore we construct reference answers by answerability analysis
$g(\cdot)$ using both the question and the realized interaction:
\begin{equation}
\vspace{-5pt}
    a_j^{*}=g(q_j,\mathcal{E},\mathcal{Z}),\qquad
    \mathcal{A}^{*}=\{a^{*}_k\}_{k=1}^{|\mathcal{Q}|}.
\end{equation}
Given $(q_j, M_{\mathcal{Z}})$, the agent produces a response with
$\hat{a}_j=f_{\pi}(q_j, M_{\mathcal{Z}})$, yielding
$\hat{\mathcal{A}}=\{\hat{a}_1,\hat{a}_2,\dots,\hat{a}_{|\mathcal{Q}|}\}$.
We compute the Environment Understanding Score (EUS) as the ratio of the number of correct answers to the total number of questions:
\begin{equation}
\vspace{-5pt}
\mathrm{EUS}
:=\frac{1}{|\mathcal{Q}|}\sum_{j=1}^{|\mathcal{Q}|}\mathbbm{I}\!\left(\hat{a}_j=a^{*}_j\right).
\end{equation}

%% file: section/4_experiment.tex
\newcommand{\bestval}[1]{\textbf{#1}}
\newcommand{\secondval}[1]{\underline{#1}}

\begin{table*}
    \centering
    \setlength{\tabcolsep}{2pt}
    \fontsize{10}{8}\selectfont
    \begin{tabular*}{0.88\textwidth}{@{\extracolsep{\fill}}ll*{7}{c}}
    \toprule
    \multirow{2}{*}{Model} & \multirow{2}{*}{Method} &
    \multirow{2}{*}{Task Score} &
    \multicolumn{6}{c}{Environment Understanding Score}
    \\
    \cmidrule(lr){4-9}
    & & &
    \multicolumn{1}{c}{Loc.} &
    \multicolumn{1}{c}{Conn.} &
    \multicolumn{1}{c}{Dir.} &
    \multicolumn{1}{c}{Match.} &
    \multicolumn{1}{c}{Prop.} &
    \multicolumn{1}{c}{Tot.}
    \\
    \midrule
    \multirow{1}{*}{GPT-5.1}
    & & \bestval{68.75} & 62.52 & 51.66 & 45.68 & 43.53 & 35.93 & 50.23 \\
    \midrule
    \multirow{4}{*}{GLM-4.6}
    & In-context & 61.16 & 62.85 & 56.40 & 53.33 & 38.24 & 35.65 & 52.41 \\
    & Mem0 & 39.29 & 55.32 & 52.84 & 50.37 & 24.71 & 28.41 & 46.21 \\
    & LangMem & 51.34 & 60.23 & 59.72 & 57.04 & 33.53 & 30.08 & 51.65 \\
    & A-MEM & 42.86 & 56.46 & 58.53 & 52.59 & 27.65 & 27.30 & 48.30 \\
    \midrule
    \multirow{4}{*}{DeepseekV3.2}
    & In-context & \secondval{63.84} & \bestval{68.41} & 59.48 & 53.09 & \bestval{49.41} & 27.58 & 54.25 \\
    & Mem0 & 43.30 & 61.87 & 60.66 & 53.58 & 44.71 & 28.97 & 52.41 \\
    & LangMem & 54.46 & \secondval{66.94} & 57.35 & 50.37 & 44.12 & 27.86 & 52.36 \\
    & A-MEM & 45.54 & \secondval{66.94} & 59.00 & 56.05 & \secondval{46.47} & 30.36 & 54.55 \\
    \midrule
    \multirow{4}{*}{Qwen3-32B}
    & In-context & 34.38 & 63.38 & \secondval{69.87} & \secondval{58.90} & 44.30 & 43.65 & \secondval{58.55} \\
    & Mem0 & 18.30 & 56.63 & 66.82 & 58.52 & 36.47 & 42.90 & 54.96 \\
    & LangMem & 27.68 & 61.37 & 67.77 & 57.28 & 37.06 & \secondval{44.01} & 56.63 \\
    & A-MEM & 18.30 & 61.05 & \bestval{71.33} & \bestval{66.67} & 44.12 & \bestval{46.52} & \bestval{60.29} \\
    \bottomrule
    \end{tabular*}
    \caption{Main experiment results. Best and second-best values in each column are highlighted in \textbf{bold} and \underline{underlined}, respectively, across all model--memory system combinations. Column abbreviations: Loc. (Location), Conn. (Connectivity), Dir. (Direction), Match (Matching), Prop. (Properties), Tot. (Overall).}
    \vspace{-10pt}
    \label{tab:main-experiments}
\end{table*}

\section{Experiment}

This section describes our experimental setup, including the evaluated models and agent baselines, followed by a discussion of the main results and associated analyses.

\subsection{Models and Agent Baselines}
We evaluate a strong proprietary model GPT-5.1~\cite{openai-gpt51} and representative open-source models (DeepSeekV3.2~\cite{deepseek32}, ChatGLM4.6~\cite{zhipu-glm46}, and Qwen3-32B~\cite{yang2025qwen3technicalreport}). On top of each open-source backbone, we compare multiple agent configurations: a naive \textit{In-context} baseline that retains the full interaction history, and memory-augmented baselines, including Mem0~\cite{mem0}, LangMem~\cite{langmem}, and A-MEM~\cite{amem} that implement different memory systems.

\vspace{-5pt}
\subsection{Main Result}
We report both task completion performance and quiz accuracy on T2QBench across models and memory mechanisms. Quiz results are further broken down by question categories (e.g., direction, location, and object properties).
Table~\ref{tab:main-experiments} reveals consistent patterns across models, memory mechanisms, and question types.

\subsection{Analysis}
\begin{wrapfigure}[11]{r}{0.35\textwidth}
    \vspace{-15pt}
    \centering
    \includegraphics[trim=0cm 0cm 0cm 0cm, clip, width=\linewidth]{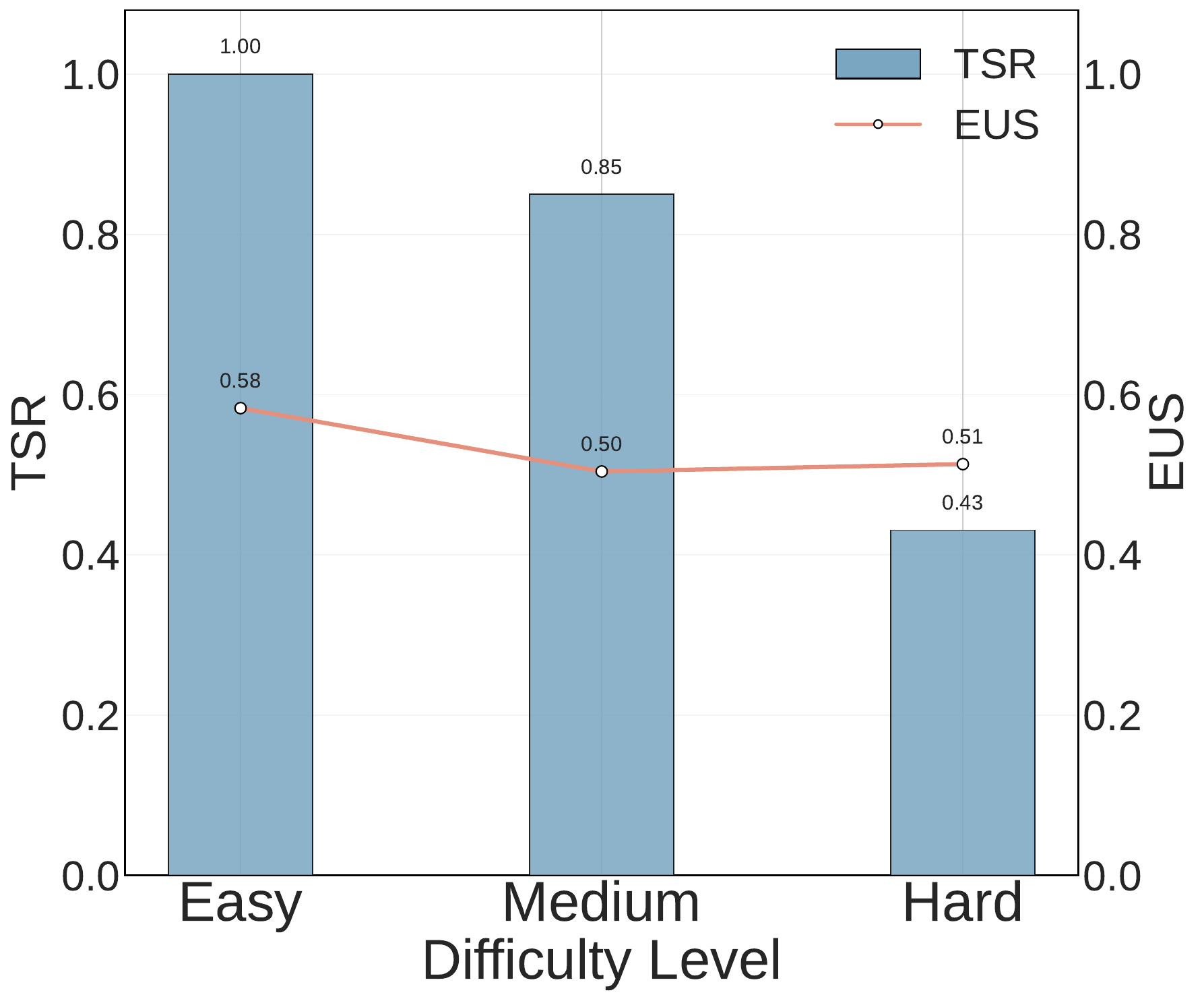}
    \vspace{-20pt}
    \caption{TSR and EUS by difficulty level of GLM-4.6 with in-context method.}
    \label{fig:different_trends_of_tsr_and_eus}
\end{wrapfigure}
In this section, we analyze the main results and figure out the answers to the research questions.
\vspace{-5pt}
\paragraph{RQ1: Different Trends of TSR and EUS Across Difficulty Levels}

To figure out the relationship between TSR and EUS, we evaluate the TSR and EUS trends across different difficulty levels. We observe that \textbf{TSR} and \textbf{EUS} exhibit markedly different behaviors as environment difficulty increases. Figure~\ref{fig:different_trends_of_tsr_and_eus} illustrates this effect for GLM-4.6 under an in-context baseline: TSR saturates at 100\% on easy environments but degrades substantially with increasing difficulty. In contrast, EUS remains comparatively stable across difficulty levels, suggesting that task success is not a reliable proxy for environment understanding.

\begin{tcolorbox}[
title=\textbf{Takeaway 4.1 for RQ1}, colback=SeaGreen!10!CornflowerBlue!10,colframe=RoyalPurple!55!Aquamarine!100!,
boxrule=0.6pt
]
\textbf{Doing is different from knowing.} Task success primarily tracks trajectory difficulty (e.g., longer solutions and tighter constraints), whereas EUS captures a more structural form of environment modeling grounded in interaction. Consequently, evaluating agents solely by completion metrics cannot fully reflect environment understanding.
\end{tcolorbox}

\paragraph{RQ2: Comparison between In-context Method and Memory Systems}
\begin{wrapfigure}{l}{0.50\textwidth}
    \vspace{-5pt}
    \centering
    \includegraphics[trim=0cm 0cm 0cm 0cm, clip, width=\linewidth]{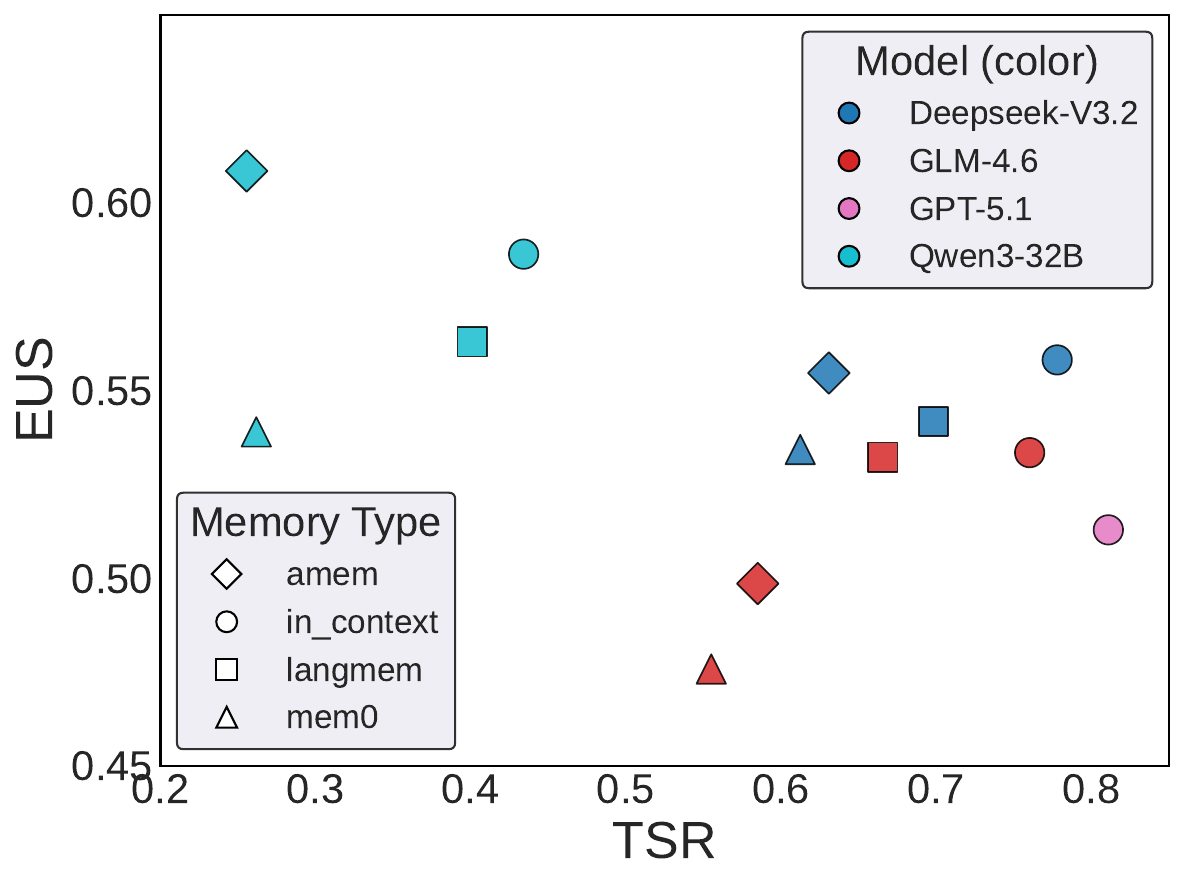}
    \vspace{-18pt}
    \caption{TSR and EUS Comparison across models and memory systems.}
    \label{fig:tsr_eus_scatter}
    \vspace{-10pt}
\end{wrapfigure}
We then examine the effect of memory systems on environment understanding. We compare a naive in-context baseline with several representative memory-augmented variants, and summarize TSR and EUS across models and memory systems in Figure~\ref{fig:tsr_eus_scatter}. Surprisingly, the in-context approach often attains the strongest overall performance in both task success and environment understanding, outperforming methods equipped with explicit memory systems. This suggests that current memory systems do not reliably help agents form a more complete understanding of exploratory environments. A plausible explanation is that, during memory construction and retrieval, high-level abstractions or summaries discard critical fine-grained evidence, resulting in worse downstream reasoning than simply retaining the full interaction context. More broadly, limited environment mastery not only lowers EUS but also harms TSR, indicating that a comprehensive understanding of the environment is also important for tasks.

\begin{tcolorbox}[
title=\textbf{Takeaway 4.2 for RQ2}, colback=SeaGreen!10!CornflowerBlue!10,colframe=RoyalPurple!55!Aquamarine!100!,
boxrule=0.6pt
]
\textbf{Current memory systems helps organization less than it hurts fidelity.} Existing memory systems tend to lose environment-specific details while still failing to distill them into structural abstractions such as layout. This gap may stem from event-centric memory systems, highlighting the need for memory mechanisms tailored to grounded world-state representation.
\end{tcolorbox}

\paragraph{RQ3: Comparison between Different Question Types}
\begin{wrapfigure}[15]{r}{0.55\textwidth}
    \vspace{-15pt}
    \centering
    \includegraphics[trim=0cm 0cm 0cm 0cm, clip, width=\linewidth]{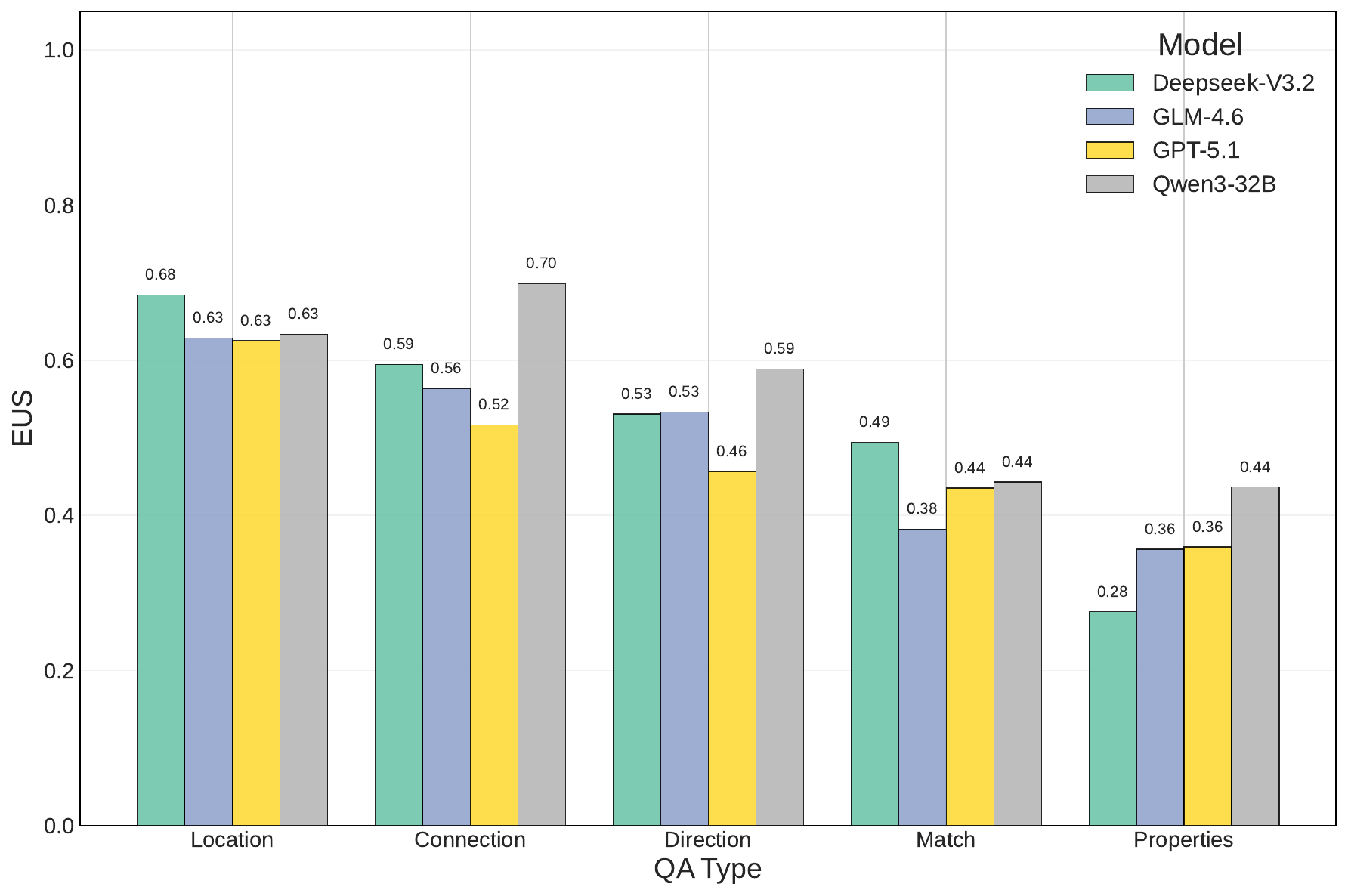}
    \vspace{-15pt}
    \caption{EUS Comparison Across Question Types: Agents have different performance on different question types.}
    \label{fig:qa_accuracy_comparison}
\end{wrapfigure}

To further investigate what hinders the agent's environment understanding, we analyze the performance of different question types in Figure~\ref{fig:qa_accuracy_comparison}. Models achieve relatively high accuracy on \textbf{location} questions, which are often directly answerable from a single recent observation. Performance drops for questions like orientation requiring multi-turn retrieval and reasoning across interactions. The largest degradation appears on questions that require \textbf{proactive} interaction to reveal latent properties or relations. Overall, these patterns suggest that insufficient exploration incentives is the dominant bottleneck than retrieval from memory.

\begin{tcolorbox}[
title=\textbf{Takeaway 4.3 for RQ3}, colback=SeaGreen!10!CornflowerBlue!10,colframe=RoyalPurple!55!Aquamarine!100!,
boxrule=0.6pt
]
\textbf{Exploration is the dominant bottleneck than retrieval.} Agents are typically optimized to complete a given task with minimal interactions, which incentivizes efficient goal-directed behavior. As a result, agents often fail to uncover environment details for deeper understanding. This suggests that the insufficient exploration incentives are the primary bottleneck than retrieval from memory.
\end{tcolorbox}

%% file: section/5_relatedwork.tex
\section{Related Work}
\subsection{LLM Agentic Benchmarks}
Agentic benchmarks increasingly evaluate LLM agents in \textbf{executable} environments that support multi-step interaction and tool use, such as realistic web navigation~\cite{zhou2023webarena,he2024webvoyager} and repository-level software engineering tasks~\cite{jimenez2023swe}. More broadly, recent suites cover apps~\cite{trivedi2024appworld}, OS or GUI-like scenarios~\cite{agashe2024agent,xie2024osworld}, and multi-environment or generalist evaluation~\cite{liu2024agentbench,mialon2023gaia}. Despite their achievements, many benchmarks remain task-centric and costly to scale, motivating more controllable and fine-grained evaluation beyond final success.
\vspace{-5pt}
\subsection{Game-based Evaluation}
Games provide dynamic, multi-skill environments that stress long-horizon planning, exploration, and state tracking. Recent benchmarks include text or conversational games~\cite{qiao2023gameeval,costarelli2024gamebench}, live computer games~\cite{hu2024gamearena,li2025llmpysc2starcraftiilearning}, and multimodal games~\cite{paglieri2024balrog}. Strategic and multi-agent settings further evaluate LLM decision-making~\cite{wu2023smartplay,duan2024gtbench,chen2024llmarena} and broader gameplay ability~\cite{huang2024far,chalamalasetti2023clembench}. Text-based adventure games are particularly attractive for grounded evaluation due to explicit state transitions and interpretable actions. Frameworks such as TextWorld~\cite{cote2019textworld} and interactive fiction resources~\cite{jericho,phan2025textquestsgoodllmstextbased} become an alternative to mirror real-world challenges. In contrast to these evaluation works, we leverage text games as a framework to conduct the entire comprehensive evaluation of the agent's environment understanding.

%% file: section/6_conclusion.tex
\vspace{-5pt}
\section{Conclusion}
\vspace{-5pt}

In this paper, we introduces \textbf{Task2Quiz(T2Q)}, a new paradigm and automatically built \textbf{T2QBench}, a benchmark spanning three difficulty levels with 30 environments, 224 coverage-oriented tasks, and 1,967 environment-grounded QA pairs. We design a unified two-stage evaluation with Task Success Rate (TSR) and Environment Understanding Score (EUS). We present a comprehensive empirical study on \textbf{T2QBench}, and reveal \textbf{three key insights}: (i) There is a gap between knowing and doing, task success rate can not reflect what agents \textbf{know} about the environment, as TSR degrades with difficulty while EUS remains comparatively stable; (ii) recent memory systems do not achieve better EUS than the naive in-context method, suggesting they are not yet effective at extracting and structuring environment-relevant information from interaction traces; and (iii) it is the lack of propensity for proactive exploration what hinders agents to form comprehensive world models.
Beyond trajectory-level, our work offers a principled route to evaluate environment modeling capability of agents, providing a diagnostic foundation for developing more robust agents with generalized abilities.

%% file: section/7_limitations.tex
\vspace{-5pt}
\section{Limitations}
\vspace{-5pt}
Despite our comprehensive analysis of agents' environment understanding, our work has several limitations.

First, due to computational constraints and the cost of API-based evaluation, we evaluate only a subset of representative open-source and closed-source models; extending coverage to a broader range of agents can improve the reliability of our findings.

Second, our benchmark is built on TextWorld framework~\cite{cote2019textworld}, whose game mechanics and maps are relatively simple compared with commercial game such as Minecraft. Future extensions could incorporate more complex dynamics, larger maps, and additional modalities. There are two possible ways to do this, both challenging: (1) adapting a commercial game as the development environment, where access to copyright-protected source code is often restricted; and (2) building a new framework from scratch using a more powerful engine such as Unity~\cite{unity_engine_2022} or Unreal Engine~\cite{unreal_engine_5}, which would require much engineering effort. Although our game framework is simple, it is sufficient to support the key insights of our work.
Finally, the world-analysis algorithm for coverage-oriented sub-tasks relies on TextWorld framework. It is implemented by a dependency tree structure and exhaustive state-space search (e.g., graph traversal over world states), which can become computationally expensive as the number of rooms and interactive objects increases for more complex games. Improving the efficiency and scalability of this algorithm remains a promising avenue for future research.

%% file: section/8_appendix.tex
\label{sec:appendix}\tcbset{
promptbox/.style={
colback=purple!10,
colframe=purple!60,
boxrule=0.8pt,
enhanced,
breakable,
listing only,
listing options={
basicstyle=\ttfamily\footnotesize,
breaklines=true,
breakatwhitespace=true
}
}
}
\section{The detail of Constructing an environment}
\label{sec:detail-of-constructing-an-environment}
We build each text-game environment with TextWorld's \texttt{GameMaker} API, which exposes the full world state by construction and allows us to precisely control layout, entities, and stateful relations. Concretely, an environment is specified through three steps: (i) defining the world topology (rooms and connectivity); (ii) instantiating entities (doors, containers, objects) and placing them; and (iii) declaring symbolic facts (e.g., locked, closed, match) that govern dynamics and answerable knowledge.

\paragraph{World topology (layout control).}
We explicitly create rooms and connect them via directional edges (e.g., \texttt{roomA.east} to \texttt{roomB.west}), yielding a controllable underlying graph. The player start location is also deterministically assigned. Since the topology is programmatically constructed, the ground-truth map (rooms and connectivity) is always available for verification and question generation.

\paragraph{Entities (object control).}
TextWorld provides typed entities that can be composed to form richer worlds. Rooms and fixed supports define where objects can appear, while doors and containers introduce stateful constraints (open, closed, locked) that shape exploration and interaction. Each entity can be named and optionally given a description used by \texttt{examine}.

\paragraph{Facts and relations (state control).}
Beyond entity placement, the environment dynamics and many evaluable properties are encoded as symbolic predicates added via \texttt{add\_fact}. This makes the world fully controllable at the level of discrete state transitions (e.g., whether a door is locked) and relational structure (e.g., which key matches which lock), and provides a clean source of environment metadata for our verifier.

\paragraph{Supported entity types.}
We use the standard types provided by \texttt{GameMaker}:
\begin{itemize}
\item \texttt{r}: room; \texttt{d}: door; \texttt{c}: container; \texttt{s}: supporter
\item \texttt{o}: portable object; \texttt{k}: key; \texttt{f}: food
\item \texttt{oven}, \texttt{stove}: specialized heat-source container or supporter
\end{itemize}

\paragraph{Supported predicates (facts).}
We mainly rely on the following predicates for controllable state and relations:
\begin{itemize}
\item \texttt{match(key, door or container)}: key--lock compatibility
\item \texttt{open(door or container)}, \texttt{closed(door or container)}, \texttt{locked(door or container)}: access state
\item \texttt{edible(food)}: consumability (optional)
\end{itemize}

\paragraph{How to add new structure (examples).}
All controllable components are added deterministically in code, e.g., (i) create rooms and connect them; (ii) create entities and place them in a room, supporter, or inventory; and (iii) attach facts to specify states and relations. For instance, a locked door and its matching key can be created by adding \texttt{locked(door)} and \texttt{match(key, door)}, while containers can be initialized with \texttt{closed(container)} before gameplay.

\paragraph{Quests and distractors (task control).}
Given a constructed world, we can (a) record a specific quest by logging an intended command sequence, or (b) automatically sample multiple random quests of bounded length. We can also inject distractor objects to increase interaction complexity without changing the underlying ground-truth world state. Together, these controls allow us to scale environment instances while keeping their structure and metadata fully known and reproducible.

\paragraph{Minimal API examples.}
Below we provide short code snippets illustrating how we control topology, entities, and symbolic facts using TextWorld's \texttt{GameMaker}.

\paragraph{Entity types.}
We use TextWorld's typed object system, e.g., \texttt{r} (room), \texttt{d} (door), \texttt{c} (container), \texttt{s} (supporter), \texttt{o} (portable object), and \texttt{k} (key).

\begin{tcolorbox}[colback=gray!3,colframe=black!50,title={Example 1: Define rooms and connectivity (topology control)}]
\begin{verbatim}
M = GameMaker()
roomA = M.new_room("Room A")
roomB = M.new_room("Room B")
M.connect(roomA.east, roomB.west)
# directed ports define the map
M.set_player(roomA)
# fix the start room
\end{verbatim}
\end{tcolorbox}

\begin{tcolorbox}[colback=gray!3,colframe=black!50,title={Example 2: Add entities and place them (entity control)}]
\begin{verbatim}
s =M.new(type="s",name="table")
roomA.add(s)
# fixed in the room
\end{verbatim}
\end{tcolorbox}

\begin{tcolorbox}[colback=gray!3,colframe=black!50,title={Example 3: Add a locked door and matching key (fact or relation control)}]
\begin{verbatim}
d=M.new_door(corridor, name="door")
M.add_fact("locked", d)
k=M.new(type="k", name="old key")
M.add_fact("match", k, d)
s.add(k)
\end{verbatim}
\end{tcolorbox}

\begin{tcolorbox}[colback=gray!3,colframe=black!50,title={Example 4: Add a container with an explicit initial state}]
\begin{verbatim}
c=M.new(type="c", name="fridge")
M.add_fact("closed", c)
  # explicit initial status
roomB.add(c)
\end{verbatim}
\end{tcolorbox}

\begin{wrapfigure}[13]{r}{0.40\textwidth}
  \vspace{-5pt}
  \centering
  \includegraphics[trim=0cm 0cm 15cm 0cm, clip, width=\linewidth, page=27]{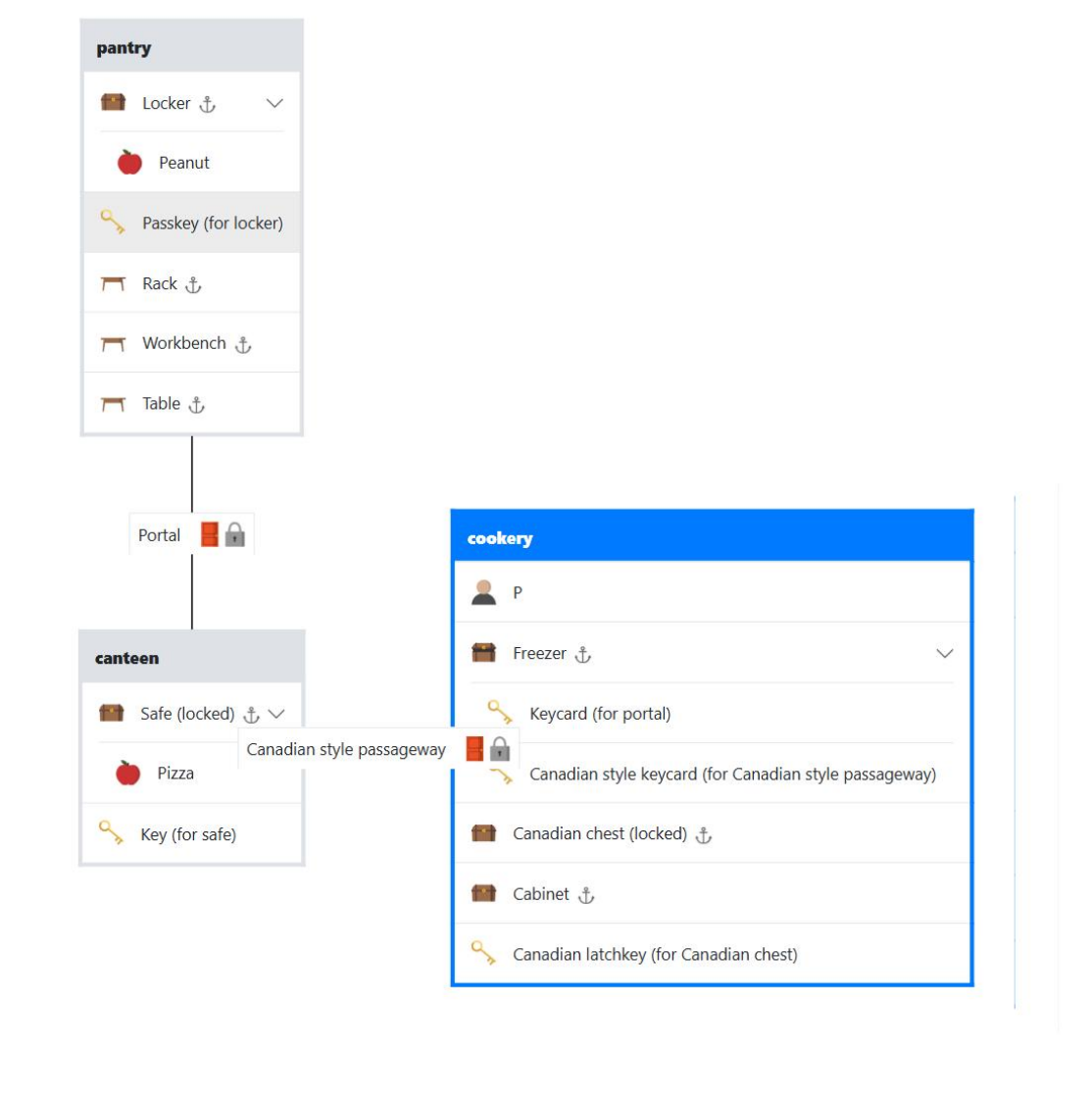}
    \vspace{-10pt}
    \caption{Example of a game environment}
    \label{fig:game_example}
    \vspace{-5pt}
  \end{wrapfigure}

\paragraph{Controllable predicates.}
We encode state and relations as symbolic facts, including \texttt{open}, \texttt{closed}, \texttt{locked}, and \texttt{match(key, door or container)}. Because these facts are added by construction, the complete environment metadata (topology, placements, and states) is fully known and can be used for deterministic verification and question generation.

\section{Example of a game environment}
Our generated game is packaged as a gym~\cite{brockman2016openai} environment. Since all gym games have only one winning state. In order to apply our task sets on one environment, we actually generate a new game environment with the same structure but different goal. So one env $\mathcal{E}$ with n tasks $\mathcal{T} = \{t_1, t_2, \dots, t_n\}$ is actually organized as n games $\mathcal{G} _{1} = (\mathcal{E}, t_1), \mathcal{G}_{2} = (\mathcal{E}, t_2), \dots, \mathcal{G}_{n} = (\mathcal{E}, t_n)$.
The game content is shown in Figure~\ref{fig:game_example}.

The task description is shown as follows:
\begin{tcolorbox}[colback=gray!3,colframe=black!50,title={Task Description}]
\begin{verbatim}
Get ready to pick stuff up
and put it in places, 
because you've just entered 
TextWorld! I hope you're 
ready to go into rooms and
interact with objects, 
Recover the key from the 
floor of the canteen.
\end{verbatim}
\end{tcolorbox}

\section{Prompt Templates}
Our prompt templates are shown as follows:
\begin{tcolorbox}[
promptbox,
title={QA Type-specific Guidelines},
colback=gray!3, colframe=black!60, boxrule=0.8pt
]
\textbf{yes\_no}: Answer ``yes'' or ``no'' only.\par
\textbf{which}: Answer with the exact choice text from the given choices.\par
\textbf{description}: Provide a concise, categorical description. Focus on object types or categories rather than specific item names. Do NOT include exits, doors, or room connections. Valid options: [object, container, supporter, food, key].\par
\textbf{where}: Answer with the location or room name.\par
\textbf{what}: Answer concisely with the specific information requested.\par
\textbf{default}: Answer concisely.
\end{tcolorbox}

\begin{tcolorbox}[
promptbox,
title={Action Prompt Template},
colback=gray!3, colframe=black!60, boxrule=0.8pt
]
Observation:\par
\{obs\}\par\medskip
Inventory:\par
\{inventory\}\par\medskip
Score: \{score\}\par\medskip
Return ONLY JSON: \{"reason": "...", "command": "<one command>"\}
\end{tcolorbox}

\begin{tcolorbox}[
promptbox,
title={QA Prompt Template},
colback=gray!3, colframe=black!60, boxrule=0.8pt
]
You answer questions about TextWorld gameplay.\par
Based on the provided context, answer the following question about the TextWorld game initial state.\par\medskip

Context:\par
\{context\}\quad (only included if context $\neq$ "")\par\medskip

\{question\}\par\medskip

Choices:\par
-- \{choice\_1\}\par
-- \{choice\_2\}\par
$\cdots$\quad (only included if choices is provided)\par\medskip

Answer should be json of the form:\par
\{\par
\quad ``answer'': ``<your answer>'',\par
\quad ``reason'': ``<why>''\par
\}\par\medskip

Answer formatting guidelines:\par
\{type\_specific\_guideline\}\par\medskip

If the question cannot be answered based on the context, set answer to ``non-answerable''.
\end{tcolorbox}

\begin{tcolorbox}[
promptbox,
title={QA Type-specific Guidelines},
colback=gray!3, colframe=black!60, boxrule=0.8pt
]
\textbf{yes\_no}: Answer ``yes'' or ``no'' only.\par
\textbf{which}: Answer with the exact choice text from the given choices.\par
\textbf{description}: Provide a concise, categorical description. Focus on object types or categories rather than specific item names. Do NOT include exits, doors, or room connections. Valid options: [object, container, supporter, food, key].\par
\textbf{where}: Answer with the location or room name.\par
\textbf{what}: Answer concisely with the specific information requested.\par
\textbf{default}: Answer concisely.
\end{tcolorbox}
\begin{figure}[H]
  \centering
  \includegraphics[trim=0cm 0cm 3cm 0cm, clip, width=0.8\linewidth, page=25]{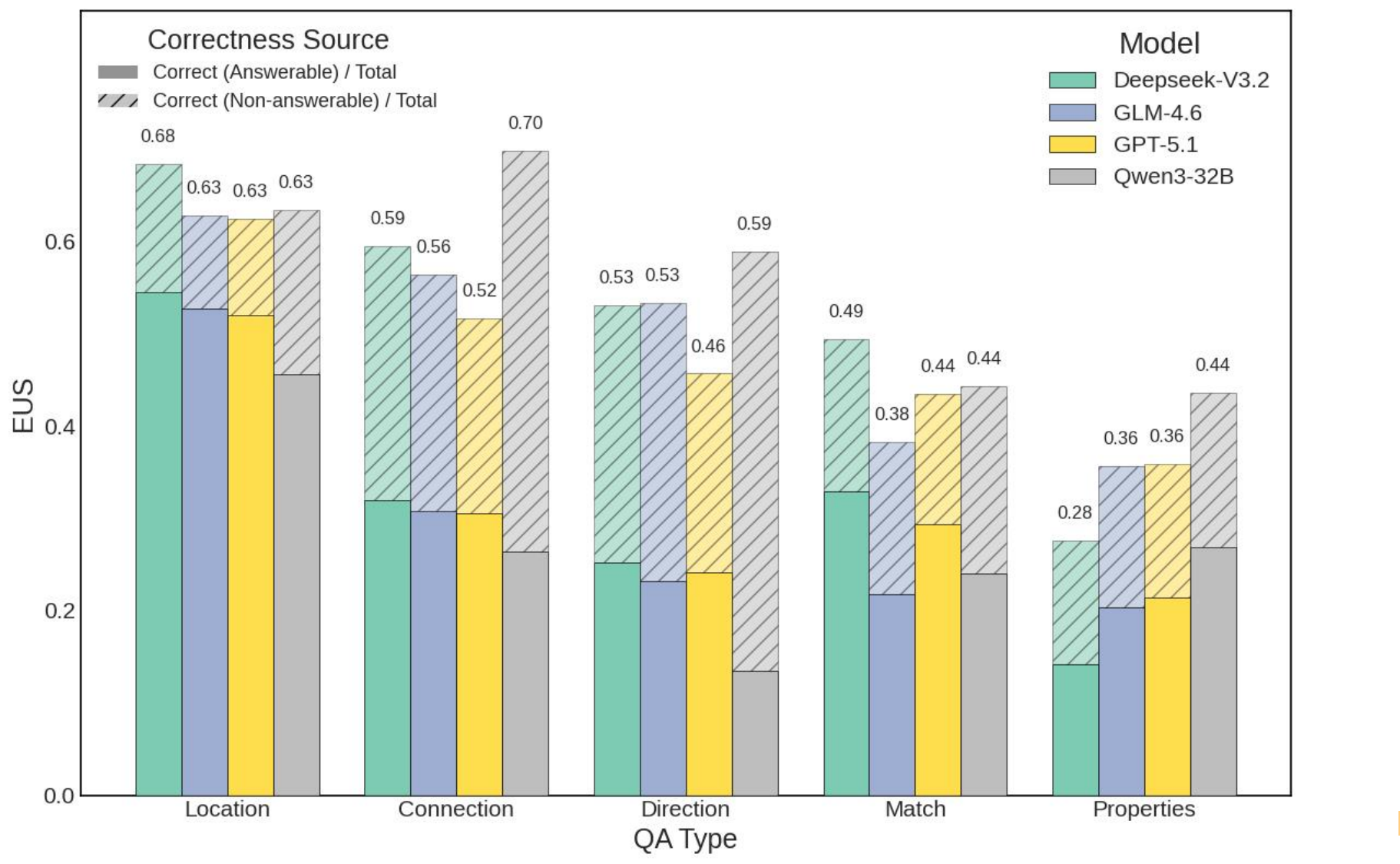}
  \caption{The comparison of answerable QA accuracy and non-answerable QA accuracy}
  \label{fig:answerable_accuracy}
  \end{figure}
\section{The comparison of answerable QA accuracy and non-answerable QA accuracy}

We compare the accuracy of answerable QA and non-answerable QA on the T2QBench dataset, as shown in Figure~\ref{fig:answerable_accuracy}. In another word, we analyze the source of correctness.
To figure out if agent can answer the exact fact ground truth, or gain accuracy simply by answering non-answerable.

From this figure, we observe that the composition of correctness sources varies substantially across models, particularly in the ratio of answerable vs.\ non-answerable QA. For location questions that can be resolved via single-turn retrieval, agents typically satisfy the prerequisite checkpoints during exploration and can answer correctly once the relevant evidence is observed. In contrast, for orientation and connectivity questions, insufficient coverage of key locations often makes the required evidence unavailable; consequently, models can frequently obtain correct outcomes by predicting \texttt{non-answerable}. For object-property questions, however, models perform poorly on both answerable and non-answerable instances, suggesting that limited proactive interaction (needed to reveal latent properties) remains a major bottleneck.